\documentclass[a4paper, 10 pt, conference]{ieeeconf}

\IEEEoverridecommandlockouts

\usepackage{amsmath,graphicx}
\usepackage[algorithm]{algorithm}
\usepackage{algorithmicx}
\usepackage[noend]{algpseudocode}
\graphicspath{ {././figures/} }
\usepackage{multirow}
\usepackage{hhline}

\usepackage{epsfig}   % for postscript graphics files
\usepackage{mathptmx} % assumes new font selection scheme installed
\usepackage{times}    % assumes new font selection scheme installed
\usepackage{amsmath}  % assumes amsmath package installed
\usepackage{amssymb}  % assumes amsmath package installed

\usepackage{array}
\usepackage{url}

\usepackage{stfloats}

\title{\LARGE \bf Driver Behavior Analysis Using Lane Departure Detection Under Challenging Conditions*
}

\author{Luis Riera$^{1}$, Koray Ozcan$^{2}$, Jennifer Merickel$^{3}$,  Mathew Rizzo$^{4}$, Soumik Sarkar$^{5}$, and Anuj Sharma$^{6}$% <-this % stops a space
\thanks{*This work was supported by Toyota Collaborative Safety Research Center and the National Institutes of Health (R01-AG017177).}% <-this % stops a space
\thanks{$^{1\&5}$Department of Mechanical Engineering, Iowa State University, Ames, IA 50011, USA
        {\tt\small lgriera@iastate.edu, soumiks@iastate.edu}}%
\thanks{$^{2\&6}$Institute for Transportation, Iowa State University, Ames, IA 50010, USA
        {\tt\small koray6@iastate.edu, anujs@iastate.edu}}%
\thanks{$^{3\&4}$Neurological Sciences, University of Nebraska Medical Center, Omaha, NE 68198, USA
        {\tt\small jennnifer.merickel@unmc.edu, matthew.rizzo@unmc.edu}}%
}

\begin{document}

\maketitle
\thispagestyle{empty}
\pagestyle{empty}

\begin{abstract}
In this paper, we present a novel model to detect lane regions and extract lane departure events (changes and incursions) from challenging, lower-resolution videos recorded with mobile cameras. Our algorithm used a Mask-RCNN based lane detection model as pre-processor. Recently, deep learning-based models provide state-of-the-art technology for object detection combined with segmentation. Among the several deep learning architectures, convolutional neural networks (CNNs) outperformed other machine learning models, especially for region proposal and object detection tasks. Recent development in object detection has been driven by the success of region proposal methods and region-based CNNs (R-CNNs). Our algorithm utilizes lane segmentation mask for detection and Fix-lag Kalman filter for tracking, rather than the usual approach of detecting lane lines from single video frames. The algorithm permits detection of driver lane departures into left or right lanes from continuous lane detections. Preliminary results show promise for robust detection of lane departure events. The overall sensitivity for lane departure events on our custom test dataset is 81.81\%.
\end{abstract}


\section{INTRODUCTION}
\label{sec:intro}

Motor vehicle collisions are a leading cause of death and disability worldwide. According to the World Health Organization, nearly 1.2 million people worldwide die and 50 million are injured every year due to traffic-related accidents. Traffic accidents result in considerable economic cost, currently estimated at ~1-2\% of average gross national product (\$518 billion globally per year)~\cite{2013Gsro}.  According to the European Accident Research and Safety Report 2013, more than 90\% of driving accidents are caused by safety-critical driver errors~ \cite{truck2013european}. Lane incursions due to driver error are a common cause of accidents. Estimates from the U.S. National Highway Traffic Safety Administration indicate that ~11\% of accidents are due to the driver inappropriately departing from their lane while traveling~\cite{doths811059}

To address this risk, Lane Departure Warning (LDW) systems are becoming a commonly deployed driver assistance technology aimed at improving on-road safety and reducing traffic accidents~\cite{narote2018review}. LDW systems typically detect lanes from low-level image features such as edges and contours. Several solutions aimed at detecting vehicle lane position and alerting drivers to potentially unsafe lane departure events have been developed~\cite{hsiao2009portable}. For example, simple image feature based systems have been developed to detect straight lines, polynomial, cubic-spline, piecewise linear, and circular arcs--relevant to lane detection~\cite{tapia2013robust}.  However, image feature based systems have predictable limitations and can become unreliable with increasing road scene complexity (e.g., shadows, low visibility, occlusions, and curves) ~\cite{kim2017fast}.  Due to these limitations, researchers have been turning their attention to machine learning (ML) based methods to overcome the above-mentioned shortcomings, most recently deep neural networks (DNN).  Region-based Convolution Neural Networks (RCNNs), a type of DNN architecture, outperform other DNN architectures in object detection and recognition applications. Due to these key advantages, we have chosen RCNNs to build a simple but robust LDW system that overcomes limitations of previous LDW systems.

The primary application of this project is to detect unsafe driver behaviors, like lane incursions or departures, in at-risk drivers with diabetes. Diabetes affects nearly 10\% of the population in the USA and continues to increase with urbanization, obesity, and aging \cite{ADA}. Drivers with diabetes have a significantly  elevated crash risk compared to the general population--presenting a pressing problem of public health and patient safety. On-road risk in diabetes is linked to disease and unsafe physiologic states (e.g., hypoglycemia). These key factors make this population a prime target for improving safety with driver assistance systems like LDW.   

This model is capable of processing large data collection representing multiple Terabytes (TBs) of video collected from at-risk drivers with diabetes. We present this lane detection model using a Mask-RCNN architecture to analyze lane departures and incursions from lane detections in challenging, lower-resolution, and noisy video recordings. Lane incursion is defined as performing an incomplete lane departure while quickly returning back to the original lane of travel. While previous literature addresses simple lane line detection, our model focuses on advancing these models by improving detection and segmentation of the driving lane area. Once the driving lane area is detected in the video frame, we tracked a centroid of convex hull region, representing the driving lane area. The centroid location with respect to the image vertical center line was used to determine if the driver was driving within the lane or s/he was denaturing from it.  Subsequently, the time series relative lane position was used to infer driving behavior.

This paper is organized as follows: \textbf{Related Works} describes previous related work done on lane departure using image-based features and machine learning based approaches. \textbf{Custom Dataset} provides general information about the data collected, annotated, and used for this project. \textbf{Proposed Model and Lane Departure} presents our approach for detecting lane departure events. Finally, the summary and discussion of our work is presented in \textbf{Conclusions.}

\section{RELATED WORKS}
\label{sec:realtivework}
Early works in lane detection and departure warning system date back to the 1990s.  Previously proposed methods in this area can be classified as low-level image feature based, machine/deep learning (DL) based approaches, or a hybrid between the two.  The most widely used LDW systems are either vision-based (e.g., histogram analysis, Hough transformation) or more recently on DL.  In general, vision-based and DL lane detection systems start by capturing images using a selected type of sensor, pre-processing the image, followed by lane line detection and tracking.  While many types of sensors have been proposed for capturing lanes images such as radars, laser range, lidar, active infrared etc., the most widely used device is a mobile camera. An alternative to vision- and DL-based systems is the use of global-positioning systems (GPS) combined with Geographic Information Systems ~\cite{KimZuwhan2008RLDa}.  However, current LDW based on GPS can be unreliable, mainly because of the often poor reliability and resolution of GPS location and speed detection, signal loss (e.g., in covered areas), and inaccurate map databases. Due to these limitations, most modern research conducted in LDW involves a utilization of Neural Networks-based solutions in some form.

Neural Networks have been a subject of investigation in the autonomous vehicles field for a while. Among the very first attempts to use a neural network for vehicle navigation, ALNINN~\cite{pomerleau1989alvinn} is considered a pioneer and one of the most influential paper.  This model is comprised of a shallow neural network that predicts actions out of captured images from a forward facing camera mounted on-board a vehicle, with few obstacles, leading to the potential use of neural networks for autonomous navigation. More recently, advances in object detection such as the contribution made by DL and Region Convolutional Neural Network (R-CNN)~\cite{girshick2014rich} in combination with Region Proposal Network (RPN)~\cite{ren2015faster} have created models such as Mask R-CNN ~\cite{he2017mask} that provide state of the art predictions.  New trends in Neural Network object detection include segmentation, which we applied in our model as an estimator for LDW.

\subsection{Image Feature Based Methods}
\label{sec:sssec:computervision}

Image feature-based lane detection is a well researched area of computer vision~\cite{NaroteSandipannP.2018Aror}. The majority of existing image-based methods use detected lane line features such as colors, gray-scale intensities, and textural information to perform edge detection. This approach is very sensitive to illumination and environmental conditions. On the Generic Obstacle and Lane Detection system proposed by Bertozzi and Broggi~\cite{bertozzi1998gold}, lane detection was done using inverse perspective mapping to remove the perspective effect and horizontal black-white-black transaction.  Their methodology was able to locate lane markings even in the presence of shadows or other artifacts in about 95\% of the situations tested. Some of the limitations to their proposed system were computational complexity, which needed well painted lane, and assumptions such as having a lane within the region of interest and fixed minimum width of lane.

In 2005, Lee and Yi~\cite{lee2005lane}  introduced the use of Sobel operator plus non-local maximum suppression (NLMS). It was built upon methods previously proposed by Lee~\cite{lee2002machine} proposing linear lane model and edge distribution function (EDF) as well as lane boundary pixel extractor (LBPE) plus Hough transform.  The model was able to overcome weak points of the EDF based lane-departure identification (LDI) system by increasing lane parameters. The LBPE improved the robustness of lane detection by minimizing missed detections and false positives (FPs) by taking advantage of linear regression analysis. Despite improvements, the model performed poorly at detecting curved lanes.

Some of the low-level image feature based models include an initial layer to normalize illumination across consecutive images, other methods rely on filters or statistic models such as random sample consensus (RANSAC)~\cite{KimZuwhan2008RLDa}.  Lately, approaches have been incorporating machine learning, more specifically, deep learning in regards to increase image quality before detection is conducted. However, image feature-based approaches require continuous lane detections and often fail to detect lanes when edges and colors are not clearly delineated (noisy), which results in inability to capture local image feature based information. End-to-end learning from deep neural networks substantially improves model robustness in the face of noisy images or roadway features by learning useful features from deeper layers of convolution.  

\begin{figure}[t]
    \includegraphics[width=0.48\textwidth]{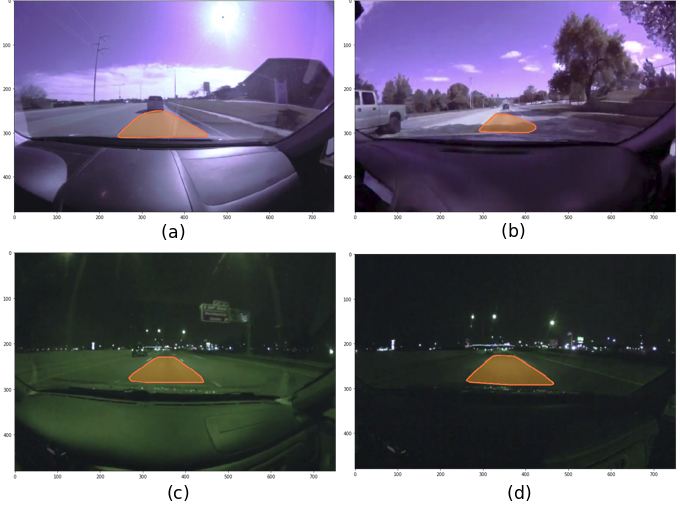}
    \caption{Examples of Challenging Lane Detections}
    \label{fig:DifficultDetections}
\end{figure}

\subsection{Deep Learning Based Methods}
\label{sec:sssec:deeplearning}
 To create lane detection models that are robust to environmental (e.g., illumination, weather) and road variation (e.g., clarity of lane markings), CNN is becoming an increasingly popular method.  Lane detection on the images shown in Fig.~\ref{fig:DifficultDetections}~(a-d) are near to impossible without using CNN. Kim and Lee~\cite{10.1007/978-3-319-12637-1_57} combined a CNN with the RANSAC algorithm to detect lanes edges on complex scenes with includes roadside trees, fences, or intersections. In their method, CNN was primarily used to enhance images.  In~\cite{huval2015empirical}, they showed how existing CNNs can be used to perform lane detection while running at frame rates required for a real-time system. Also, Ozcan et al.~\cite{ozcan2017traffic} discussed how they overcame the difficulties of detecting traffic signs from low-quality noisy videos using chain-code aggregated channel features (ACF)-based model and a CNN model, more specifically Fast-RCNN.

More recently, in~\cite{BeiHe2016Aarl}, they used a Dual-View Convolutional Neural Network (DVCNN) with hat-like filter and optimized simultaneously the frontal-view and the top-view cameras. The hat-like filter extracts all potential lane line candidates, thus removing most of FPs. With the front-view camera, FPs such as moving vehicles, barriers, and curbs were excluded. Within the top-view image, structures other than lane lines such as ground arrows and words were also removed.

%--------------------------------------------------------------------------------------------------------------------------------

\begin{figure}[t]
    \includegraphics[width=0.48\textwidth]{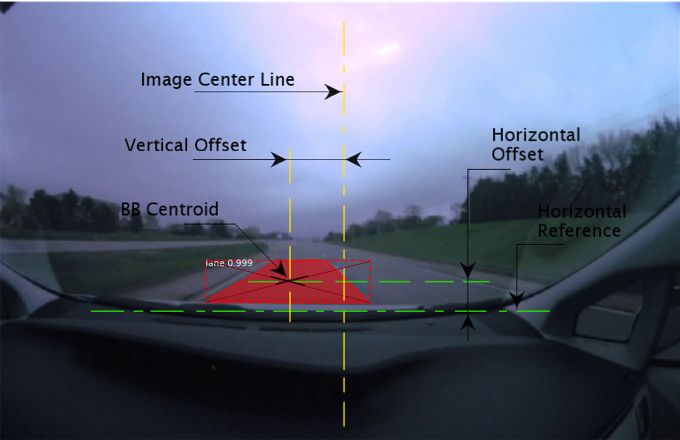}
    \caption{Lane Mask Centroid Offset Diagram}
    \label{fig:MaskCentroidOffset}
\end{figure}

%--------------------------------------------------------------------------------------------------------------------------------

\begin{figure*}[b]
    \centering
    \includegraphics[width=15.0cm]{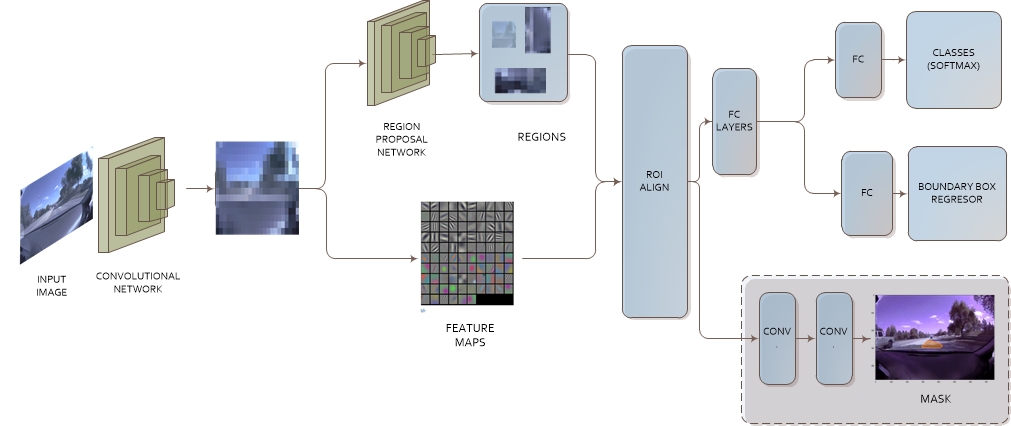}
    \caption{Segmentation with Mask-RCNN}
    \label{fig:Segmentation_Mask-RCNN}
\end{figure*}

%------------------------------------------------------------------------------------------------------------
\subsection{Lane departure models}
\label{sec:sssec:landedepartureincursion}
% check A Learning-Based Approach for Lane Departure Warning Systems with a Personalized Driver Model

The objective of Lane Departure Prediction (LDP) is to predict if the driver is likely to leave the lane with the goal of warning drivers in advance of the lane departure so that they may correct the error before it occurs (avoiding a potential collision). This improves on LDW systems, which simply alert the driver to the error after it has occurred. LDP algorithms can be classified into one of the following three categories: vehicle-variable-based, vehicle-position estimation, and detection of the lane boundary using real-time captured road images.  They all use real-time captured images~\cite{wang2018learning}.

The TLC model has been extensively used on production vehicles \cite{gaikwad2015lane}. TLC systems evaluate the lane and vehicle state relying on vision-based equipment and perform TLC calculations online using a variety of algorithms.  A TLC threshold is used to trigger an alert to the driver. Different computational methods are used with regard to the road geometries and vehicle types. Among these methods, the most common method used is to predict the road boundary, the vehicle trajectory, and then calculate intersection time of the two at the current driving speed. On small curvature roads, the TLC can be computed as the ratio of lateral distance to lateral velocity or the ratio of the distance to the line crossing. \cite{mammar2006time} Studies suggest that TLC  tend to have a higher false alarm rate (FAR) when the vehicle is driven close to lane boundary~\cite{wang2018learning, mammar2006time}.

Wang et al.~\cite{wang2018learning} proposed a online learning-based approach to predict unintended lane-departure behaviors (LDB) depending on personalized driver model (PDM) and Hidden Markov Model (HMM).  The PDM describes the driver’s lane-keeping and lane-departure behaviors by using a joint-probability density distribution of Gaussian mixture model (GMM) between vehicle speed, relative yaw angle, relative yaw rate, lateral displacement, and road curvature. PDM  can discern the characteristics of individual’s driving style. In combination with HMM to estimate the vehicle’s lateral displacement, they were able to reduce the FAR by 3.07. 

%------------------------------------------------------------------------------------------------------------
\section{Custom Dataset}
\label{sec:Dataset}
Our dataset is was collected as part of a clinical study where 77 legally licensed and active older drivers  (ages 65-90, µ=75.7; 36 female, 41 male) were recruited.  The aim of that particular project is to study the driving behavior of individuals with disabilities condition. Drivers who had physical limitations were permitted if they met state licensure standards as these limitations are ubiquitous in older adults.  Each driver drove in their typical environment with their typical strategies and driving behaviors for 3-months (total data collection embody nearly 19.3 years).  One of our contribution to this study was the detection of lane departure and incursion. For this task, we used 4,162 annotated images to train our model.  The images had a resolution of 752x480 and the videos run at an average of 25 fps.  These images were split into Training (70\%), Validation (15\%), and Test (15\%) sets. Amount all videos, we tested on our lane crossing/departure algorithm in 30 diverse videos.

%------------------------------------------------------------------------------------------------------------
\section{Proposed Model and Lane Departure\\}
\label{sec:LaneDepartureFastRCNN}
Lane detection in presence of noisy, lower-resolution image data presents significant challenges. Illumination, color contrasts, and image resolution immediately prohibit the use of low-level image feature-based algorithms for detecting the lanes. Consequently, we turned our attention to machine/DL based models to detect lane regions as these models perform better than low-level image feature-based algorithms for given lower quality recordings on custom dataset. We selected Mask-RCNN~\cite{matterport_maskrcnn_2017} architecture since we were mainly interested in segmented lane regions within the image and we could tolerate 5 \textit{fps} \cite{he2017mask}, while it provide a state-of-the-art \textit{m}AP (mean average precision).  The Mask-RCNN architecture, illustrated in Fig.  \ref{fig:Segmentation_Mask-RCNN}, can be divided into two networks. The first network is the  region proposal network (RPN) used for generating region proposals and a second network that use these proposals to detect objects. Video processing pipeline including detection and tracking is given in Algo.~\ref{alg:VideoControl}.

% As LDW systems are meant to be real-time applications, current single-pass architectures such as YOLO (You Only Look Once) and SSD  (Single Shot Detector) must meet a minimum threshold of 25 frames per second (\textit{fps}) to be applicable to real-time use.  However, while our first attempt was to use YOLO v3, we were not able to train a model due to the quality of images we have been working with. Consequently, we turned our attention to more robust architectures, particularly to the RCNN.  

% The train model used parameters shown in Tab. \ref{tab:MaskRCNN_hyperparameters}
% https://www.overleaf.com/project/5c5cc51ac5ea8e0e4811dd0b

% \begin{table}
% \renewcommand{\arraystretch}{1.4}
% \caption{Mask-RCNN Hyperparameters}
% \label{tab:MaskRCNN_hyperparameters}
% \centering

% \begin{tabular}{|l|r|}
% \hline
%     \textbf{Hyperparameter} & \textbf{Value}  \\
%     \hline
%     Backbone  &  ResNet-50 \\
%     Gradient Clip Norm & 5.0\\
%     Learning Rate & 0.001\\
%     Learning Momentum & 0.9\\
%     Weight decay & 0.0001\\
%     RPN Anchors per Image & 256\\
% \hline
% \end{tabular}
% \end{table}

%------------------------------------------------------------------------------------------------------------
\begin{algorithm}
\caption{Video Control Algorithm}\label{alg:VideoControl}
\begin{algorithmic}

\Procedure{VideoCapture}{$video$} \Comment{mp4}
    \State $frame\gets video$
        \While{$frame\not=Null$}\Comment{Loop until video end}
            \State $M\gets detection$\Comment{Mask}
            \State $Display\gets Tracking(M)$
            \State $frame\gets video$
        \EndWhile\label{vidoe_controld_endwhile}
\EndProcedure
\end{algorithmic}
\end{algorithm}

%------------------------------------------------------------------------------------------------------------

\begin{figure*}[t]
    \centering
    \includegraphics[width=17.5cm]{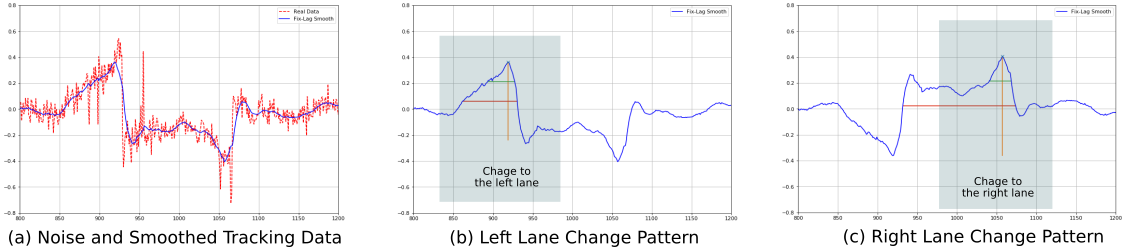}
    \caption{Synopsis of a lane change using Mask Centroid Tracking}
    \label{fig:LaneChange}
\end{figure*}

%------------------------------------------------------------------------------------------------------------
\subsection{Lane Detection}
Our Mask-RCNN based model was configured using ResNet-50 as backbone with a learning rate of 0.001, a learning momentum of 0.9, and 256 RPN Anchors per image. It was trained to detect lane regions only, using segmented mask, on the contrary to other lane detection models where their main goal is to detect the lane lines. Lane detection in lieu to lines detection was considerably easier given the quality of the images we were working with. This approach provided us a lane segmentation mask, which was later used to track the lane regions. To mitigate FPs, we used a Region of Interest (ROI) skim mask that concealed areas not relevant to our view of interest. Fig.~\ref{fig:DifficultDetections}~(a-d) provides some of the example detections during daytime, nighttime, and shadowy conditions on the road.

%-----------------------------------------------------------------------------------------------------------
\subsection{Lane Tracking}
\label{sec:sssec:LandeTracking}
Mask tracking algorithm used for lane departure and incursion predictions as explained in~Algo.~\ref{alg:LaneTracking}. Once the lane mask regions were detected, the point coordinates conforming the mask were used to compute a convex hull enclosing the mask.  For this purpose, we employed a Quickhull algorithm, which is shown in Algo.~\ref{alg:ConvexHullAlgorithm}, in order to obtain a Convex Hull polygon.  Next, a centroid of the convex hull was calculated. Our model used the centroid in order to track its vertical and horizontal offset of the vehicle within the lane as shown in Fig.~\ref{fig:MaskCentroidOffset}. For reference, the vertical offset was calculated according to an imaginary vertical line in the middle of the image as illustrated in Fig.~\ref{fig:MaskCentroidOffset}. Also, the horizontal reference was chosen to be a imaginary line between the vehicle and the detected mask. The horizontal offset was not used in our quest; however, it was implemented to detect driving separation distance possibly useful for acceleration and braking.

The offsets were calculated using the distance between a line and a point in 2D space Eq.~\eqref{distance_line2point}.  The offset units were measured in number of pixels. These offsets were first tracked over time, then normalized by their means, centered at zero and smoothed using a Fix-lag Kalman filter as it is shown in Fig.~\ref{fig:LaneChange} (a).  We found that centering around zero allows us better generalization among drivers since cameras are not necessarily mounted on the same location among different vehicles.

\begin{equation}\label{distance_line2point}{distance}(ax+by+c=0, (x_0, y_0)) = \frac{|ax_0+by_0+c|}{\sqrt{a^2+b^2}} \end{equation}

%------------------------------------------------------------------------------------------------------
\subsection{Lane departure classification}
\label{sec:sssec:DepartureClassifier}
The plots in Fig.\ref{fig:LaneChange} (b) and (c)  illustrate the typical patterns observed when the line lanes are crossed towards left or right, respectively.  These two plots were obtained while a driver changes from the right lane to the left lane and back to the right lane. It is clear that there is a high peak that starts developing as the driver departs from the center of its lane following by depression zone and a trend to go back to zero. Detecting and measuring this pattern is the core idea of our algorithm in~\ref{alg:LaneTracking} that predicts the type of lane crossing that has occurred.  Our test dataset had a limited sample of verified incursion events (N=3). Based on available experiments on lane incursion, vertical offset showed a comparable pattern to lane departures with key differences in peak shallowness and absence of a depression zone. That is how the lane incursion events are ditinguished from lane departures in~\ref{alg:LaneTracking}.

%----------------------------------------------------------------------------------------------------------

\begin{algorithm}[t]
\caption{Mask Tracking Algorithm}\label{alg:LaneTracking}
\begin{algorithmic}

\Procedure{Tracking}{$M$} \Comment{Mask}

\State $M_{centroid} \gets M_{convexHall}\gets M$
% \State $M_{centroid} \gets M_{convexHall}$
% \State $M_{norm} \gets Norm(M_{centroid})$
\State $M_{smooth} \gets series(M_{norm}) \gets Norm(M_{centroid})$
\State $prediction \gets classifier(M_{smooth})$
\State $return \gets prediction $
\EndProcedure

\Function{classifier}{$M_{smoothing,}$} \Comment{smooth series}
\State $max \gets Constant $ \Comment{Max distance between peaks}
\State $lane_{change}\gets (None, None) $
\State $lane_{incursion}\gets (None, None) $

\State $M_{mirror} \gets Mirror(M_{smooth})$

\State $orig_{peak} \gets M_{smooth}$
\State $mirr_{peak} \gets Mirror(M_{mirror})$

\If{$length(orig_{peak})>0 ~ \& ~ length(mirr_{peak})>0$}
    \If{$|orig_{peak}- mirr_{peak}|<max$}
        \If{$0<orig_{peak}-mirr_{peak}<max$}
            \State $lane_{change} \gets ('left', orig_{peak})$
            \State $return \gets (lane_{change}, lane_{incursion})$
        \ElsIf{$0<mirr_{peak}-orig_{peak}<max$}
            \State $lane_{change} \gets ('right', mirr_{peak})$
            \State $return \gets (lane_{change}, lane_{incursion})$
        \EndIf
    \Else
        \State $return \gets (lane_{change},('incursion', orig_{peak}))$
    \EndIf
\EndIf
\State $return \gets (lane_{change}, lane_{incursion}) $ \Comment{no change}

\EndFunction

\Function{Mirror}{$M_{smooth}$} \Comment{smooth series}
    \For{$X~in~M_{smooth}$}
        \State $M_{mirror} \gets -X $
    \EndFor
    \State $return \gets M_{mirror}$
\EndFunction

\end{algorithmic}
\end{algorithm}

\section{Results}
\label{sec:results}

\subsection{Testing of the Lane Detection Model}
\label{sec:sssec:validation}

Our main performance indicator for the trained Mask-RCNN model was the \textit{m}AP calculate according to Eq.~\eqref{mAP_EQ}, where TP, FP, TN, and FN denote the true positive, false positive, true negative and false negative, respectively with an intersection over union (IoU) criteria, where the IoU is calculated as per Eq.~\eqref{IoU}. The thresholds for in Eq.~\eqref{mAP_EQ} was set at 0.5, this means that any predicted object is considered a TP if its IoU with respect to the ground trust is greater than 0.5. The overall \textit{m}AP was calculated to be 0.82 for lane detections on test dataset.

\begin{equation}\label{IoU}{IoU(A,B)} = \frac{A \cap B}{A \cup B} \end{equation}

\begin{equation}\label{mAP_EQ}{\textit{m}AP} = \frac{1}{|thresholds|} \sum_{t} \frac{TP(t)}{TP(t)+FP(t)+FN(t)} \end{equation}

%------------------------------------------------------------------------------------------------------------

% https://en.wikipedia.org/wiki/Quickhull
\begin{algorithm}[t]
\caption{2D QuickHull}\label{alg:ConvexHullAlgorithm}
\begin{algorithmic}
\State $Input \gets a ~set~ S~ of n~ points$ \Comment{at least 2 points in S}
\Procedure{QuickHull}{$S$} \Comment{Gets convex hull from S }

\State    $Convex Hull \gets {\{\}} $
\For{$each~point~in~S $}
    \State  $A \gets left~ most~ point$
    \State  $B \gets right~ most~ point$
    \State  $S1 \gets points~ in~ S~ right~ to~ oriented~ line~ AB$
    \State  $S2 \gets points~ in~ S~ right~ to~ oriented~ line~ BA$
\EndFor
\State $FindHull (S1, A, B) $
\State $FindHull (S2, B, A) $
\EndProcedure

\Function{FindHull}{$Sk, P, Q$} \Comment{get points right of P to Q}
\For{$each~point~in~Sk $}
    \State $C \gets find~ farthest~ point~ from ~PQ$
    \State $S0 \gets points~ inside~ triangle~ PCQ$
    \State $S1 \gets points~ right~ side~ line~ PC$
    \State $S2 \gets points~ right~ side~ line~ CQ$
\EndFor
 
\State $FindHull(S1, P, C)$
\State $FindHull(S2, C, Q)$
\EndFunction

\State $Output \gets Convex Hull$
\end{algorithmic}
\end{algorithm}
%-----------------------------------------------------------------------------------------------------
\subsection{Lane Crossing Algorithm Results}
\label{sec:sssec:results}
We tested our lane departure algorithm using 30 short driving videos from 1 to 3 minutes long each. Diversity of drivers and environmental conditions were considered when selecting the videos. Each of the videos had at least one occurrance of lane crossing and in some circumstances, one line of the lane or portion of it was not visible(i.e. lane bifurcation on a highway exit). We used the algorithm to test for lane changes to the left, right, and lane incursion, the results are summarized in Table \ref{tab:results}.  While we did not have a good number of representative videos with lane incursions, we were able to detect two out of three incursions. TP corresponds lane departure events towards left or right that are classified correctly whereas TNs are events when vehicles stays in lane and the algorithm does not detect any lane departures. FP corresponds to incorrect lane departure detections whereas FNs are missed lane departure events. Sensitivity is defined as the ratio of the number of TP to the summation of TP and FN. Overall sensitivity is calculated to be 0.8181. It is observed that the model is very susceptible to the offset noise during the lane detection and the parameters used for peak detection algorithm.  While we mitigated the noise using smoothing techniques, a robust lane detection model is essential to increase the efficiency of the algorithm.

% Specificity, on the other hand, is defined as the ratio of TN to the summation of TN and FPs.
%-----------------------------------------------------------------------------------------------------
\begin{table}[t]
\renewcommand{\arraystretch}{1.4}
\caption{Lane Change Algorithm Test Results}
\label{tab:results}
\centering

\begin{tabular}{|c||c|c|c|c|c|c|c|c|}
\hline
\multirow{2}{*}{Videos} & 
    \multicolumn{2}{|c|}{Crossing} & 
    \multicolumn{2}{|c|}{TP} & 
    \multicolumn{2}{|c|}{FP} & 
    \multicolumn{2}{|c|}{FN} \\
\hhline{~--------}    
~  & L & R & L & R & L & R & L & R  \\
\hline
\hline
30  &  11  &  22  &  9  &  18  &  4  &  11  &  2  &  4 \\
\hline
\hline
\end{tabular}
\end{table}

\section{Conclusion}
\label{sec:conclusion}
We proposed a novel algorithm to detect and differentiate lane departure events, including incursions, on lower-resolution video recordings with challenging conditions. In our novel implementation, the model was trained to detect lanes departures with a sensitivity of ~0.82. Future investigations will expand our model to a wider variety of vehicle classes, which will likely improve FP rates, and use of segmented masks to detect lane types and improve lane incursion detection.  An area that should be further explored is the use of horizontal offset as a mean to detect proximity even when image perspectives are subject to chirp effect.  While our implementation was performed using only pre-recorded videos, utilizing a convex hull centroid offset may permit lane tracking during real-time implementation on vehicles. Our results underscore the feasibility and utility of applying DL models to autonomous driving systems, LDW/LDP, advanced driver assistance systems, and on-road interventions to improve safety in medically at-risk populations.

\section*{ACKNOWLEDGMENT}
We thank Michelle Nutting for her diligence and commitment in annotating our video data and the Mind \& Brain Health Labs at the University of Nebraska Medical Center's for coordinating this study's data collection.

\bibliographystyle{IEEEtran}
\bibliography{ITCS19}

\begin{thebibliography}{10}
\providecommand{\url}[1]{#1}
\csname url@rmstyle\endcsname
\providecommand{\newblock}{\relax}
\providecommand{\bibinfo}[2]{#2}
\providecommand\BIBentrySTDinterwordspacing{\spaceskip=0pt\relax}
\providecommand\BIBentryALTinterwordstretchfactor{4}
\providecommand\BIBentryALTinterwordspacing{\spaceskip=\fontdimen2\font plus
\BIBentryALTinterwordstretchfactor\fontdimen3\font minus
  \fontdimen4\font\relax}
\providecommand\BIBforeignlanguage[2]{{%
\expandafter\ifx\csname l@#1\endcsname\relax
\typeout{** WARNING: IEEEtran.bst: No hyphenation pattern has been}%
\typeout{** loaded for the language `#1'. Using the pattern for}%
\typeout{** the default language instead.}%
\else
\language=\csname l@#1\endcsname
\fi
#2}}

\bibitem{2013Gsro}
\emph{\BIBforeignlanguage{eng}{Global status report on road safety 2013 :
  supporting a decade of action.}}\hskip 1em plus 0.5em minus 0.4em\relax
  Geneva, Switzerland: World Health Organization, 2013.

\bibitem{truck2013european}
V.~Truck, ``European accident research and safety report 2013,''
  \emph{Gothenburg, January}, 2013.

\bibitem{doths811059}
U.~D. of~Transportation, ``National motor vehicle crash causation survey, Tech.
  Rep. DOT HS 811 059, 2008.

\bibitem{narote2018review}
S.~P. Narote, P.~N. Bhujbal, A.~S. Narote, and D.~M. Dhane, ``A review of
  recent advances in lane detection and departure warning system,''
  \emph{Pattern Recognition}, vol.~73, pp. 216--234, 2018.

\bibitem{hsiao2009portable}
P.-Y. Hsiao, C.-W. Yeh, S.-S. Huang, and L.-C. Fu, ``A portable vision-based
  real-time lane departure warning system: day and night,'' \emph{IEEE
  Transactions on Vehicular Technology}, vol.~58, no.~4, pp. 2089--2094, 2009.

\bibitem{tapia2013robust}
R.~Tapia-Espinoza and M.~Torres-Torriti, ``Robust lane sensing and departure
  warning under shadows and occlusions,'' \emph{Sensors}, vol.~13, no.~3, pp.
  3270--3298, 2013.

\bibitem{kim2017fast}
J.~Kim, J.~Kim, G.-J. Jang, and M.~Lee, ``Fast learning method for
  convolutional neural networks using extreme learning machine and its
  application to lane detection,'' \emph{Neural Networks}, vol.~87, pp.
  109--121, 2017.

\bibitem{ADA}
\BIBentryALTinterwordspacing
A.~D. Association. (2018) Statistics about diabetes. [Online]. Available:
  \url{http://www.diabetes.org/diabetes-basics/statistics/}
\BIBentrySTDinterwordspacing

\bibitem{KimZuwhan2008RLDa}
Z.~Kim, ``Robust lane detection and tracking in challenging scenarios,''
  \emph{IEEE Transactions on Intelligent Transportation Systems}, vol.~9,
  no.~1, pp. 16--26, 2008.

\bibitem{pomerleau1989alvinn}
D.~A. Pomerleau, ``Alvinn: An autonomous land vehicle in a neural network,'' in
  \emph{Advances in neural information processing systems}, 1989, pp. 305--313.

\bibitem{girshick2014rich}
R.~Girshick, J.~Donahue, T.~Darrell, and J.~Malik, ``Rich feature hierarchies
  for accurate object detection and semantic segmentation,'' in
  \emph{Proceedings of the IEEE conference on computer vision and pattern
  recognition}, 2014, pp. 580--587.

\bibitem{ren2015faster}
S.~Ren, K.~He, R.~Girshick, and J.~Sun, ``Faster r-cnn: Towards real-time
  object detection with region proposal networks,'' in \emph{Advances in neural
  information processing systems}, 2015, pp. 91--99.

\bibitem{he2017mask}
K.~He, G.~Gkioxari, P.~Doll{\'a}r, and R.~Girshick, ``Mask r-cnn,'' in
  \emph{Proceedings of the IEEE international conference on computer vision},
  2017, pp. 2961--2969.

\bibitem{NaroteSandipannP.2018Aror}
S.~P. Narote, P.~N. Bhujbal, A.~S. Narote, and D.~M. Dhane,
  ``\BIBforeignlanguage{eng}{A review of recent advances in lane detection and
  departure warning system},'' \emph{\BIBforeignlanguage{eng}{Pattern
  Recognition}}, vol.~73, pp. 216--234, 2018.

\bibitem{bertozzi1998gold}
M.~Bertozzi and A.~Broggi, ``Gold: A parallel real-time stereo vision system
  for generic obstacle and lane detection,'' \emph{IEEE transactions on image
  processing}, vol.~7, no.~1, pp. 62--81, 1998.

\bibitem{lee2005lane}
J.~W. Lee and U.~K. Yi, ``A lane-departure identification based on lbpe, hough
  transform, and linear regression,'' \emph{Computer Vision and Image
  Understanding}, vol.~99, no.~3, pp. 359--383, 2005.

\bibitem{lee2002machine}
J.~W. Lee, ``A machine vision system for lane-departure detection,''
  \emph{Computer vision and image understanding}, vol.~86, no.~1, pp. 52--78,
  2002.

\bibitem{10.1007/978-3-319-12637-1_57}
J.~Kim and M.~Lee, ``Robust lane detection based on convolutional neural
  network and random sample consensus,'' in \emph{Neural Information
  Processing}, C.~K. Loo, K.~S. Yap, K.~W. Wong, A.~Teoh, and K.~Huang,
  Eds.\hskip 1em plus 0.5em minus 0.4em\relax Cham: Springer International
  Publishing, 2014, pp. 454--461.

\bibitem{huval2015empirical}
B.~Huval, T.~Wang, S.~Tandon, J.~Kiske, W.~Song, J.~Pazhayampallil,
  M.~Andriluka, P.~Rajpurkar, T.~Migimatsu, R.~Cheng-Yue, \emph{et~al.}, ``An
  empirical evaluation of deep learning on highway driving,'' \emph{arXiv
  preprint arXiv:1504.01716}, 2015.

\bibitem{ozcan2017traffic}
K.~Ozcan, S.~Velipasalar, and A.~Sharma, ``Traffic sign detection from
  lower-quality and noisy mobile videos,'' in \emph{Proceedings of the 11th
  International Conference on Distributed Smart Cameras}.\hskip 1em plus 0.5em
  minus 0.4em\relax ACM, 2017, pp. 15--20.

\bibitem{BeiHe2016Aarl}
B.~He, R.~Ai, Y.~Yan, and X.~Lang, ``Accurate and robust lane detection based
  on dual-view convolutional neutral network,'' in \emph{2016 IEEE Intelligent
  Vehicles Symposium (IV)}, ser. 2016 IEEE Intelligent Vehicles Symposium
  (IV).\hskip 1em plus 0.5em minus 0.4em\relax IEEE, 2016, pp. 1041--1046.

\bibitem{wang2018learning}
W.~Wang, D.~Zhao, W.~Han, and J.~Xi, ``A learning-based approach for lane
  departure warning systems with a personalized driver model,'' \emph{IEEE
  Transactions on Vehicular Technology}, vol.~67, no.~10, pp. 9145--9157, 2018.

\bibitem{gaikwad2015lane}
V.~Gaikwad and S.~Lokhande, ``Lane departure identification for advanced driver
  assistance,'' \emph{IEEE Transactions on Intelligent Transportation Systems},
  vol.~16, no.~2, pp. 910--918, 2015.

\bibitem{mammar2006time}
S.~Mammar, S.~Glaser, and M.~Netto, ``Time to line crossing for lane departure
  avoidance: A theoretical study and an experimental setting,'' \emph{IEEE
  Transactions on Intelligent Transportation Systems}, vol.~7, no.~2, pp.
  226--241, 2006.

\bibitem{matterport_maskrcnn_2017}
W.~Abdulla, ``Mask r-cnn for object detection and instance segmentation on
  keras and tensorflow,'' \url{https://github.com/matterport/Mask_RCNN}, 2017.

\end{thebibliography}

\end{document}